\documentclass[a4paper,twoside]{article}
\usepackage{times}
\usepackage{graphicx}
\usepackage{titlesec}
\usepackage{epsfig}
\usepackage{amstext}
\usepackage{amsmath}
\usepackage{amssymb}
\usepackage{subcaption}
\usepackage{dblfloatfix}%
\usepackage{lipsum}
\usepackage{booktabs}
\usepackage{tabularx}
\usepackage{colortbl}
\usepackage{multirow}
\usepackage{multicol}
\usepackage{comment}
\usepackage{xfrac}
\usepackage{hyphenat}
\usepackage{blindtext}
\newcommand{\degree}[1]{${#1}^o$}
\def\360{\degree{360}}
\usepackage{calrsfs}
\DeclareMathAlphabet{\pazocal}{OMS}{zplm}{m}{n}

\newcommand{\feats}{\pazocal{X}}

\newcommand{\mask}{\pazocal{M}}
\newcommand{\im}{\pazocal{I}}

\newcommand{\loss}{\pazocal{L}}   
\newcommand{\Attention}{\pazocal{A}}
\newcommand{\vgg}{\Phi}

\usepackage[pagebackref=true,breaklinks=true,colorlinks,bookmarks=false]{hyperref}

\usepackage{pslatex}
\usepackage{apalike}
\usepackage{SCITEPRESS}     

\begin{document}

\title{Towards Full-to-Empty Room Generation with Structure-Aware Feature Encoding and Soft Semantic Region-Adaptive Normalization}

\author{Vasileios Gkitsas,\,  Nikolaos Zioulis,\,  Vladimiros Sterzentsenko,\, Alexandros Doumanoglou,\,  Dimitrios Zarpalas \\
\affiliation{Centre for Research and Technology Hellas, Thessaloniki, Greece}
}
\keywords{Deep Learning, Omnidirectional Vision, Image-to-Image translation, Depth Estimation}



\abstract{The task of transforming a furnished room image into a background-only is extremely challenging since it requires making large changes regarding the scene context while still preserving the overall layout and style. In order to acquire photo-realistic and structural consistent background, existing deep learning methods either employ image inpainting approaches or incorporate the learning of the scene layout as an individual task and leverage it later in a not fully differentiable semantic region-adaptive normalization module. To tackle these drawbacks, we treat scene layout generation as a feature linear transformation problem and propose a simple yet effective adjusted fully differentiable soft semantic region-adaptive normalization module (softSEAN) block. We showcase the applicability in diminished reality and depth estimation tasks, where our approach besides the advantages of mitigating training complexity and non-differentiability issues, surpasses the compared methods both quantitatively and qualitatively. Our softSEAN block can be used as a drop-in module for existing discriminative and generative models. Implementation is available on \href{https://vcl3d.github.io/PanoDR//}{vcl3d.github.io/PanoDR/}.}

\maketitle
\section{Introduction}
In recent years, deep learning has witnessed an unprecedented pace of improvement, most notably concerning the generation of high-dimensional content. Contemporary approaches that leverage generative adversarial networks \cite{goodfellow2014generative} have shown impressive achievements in generating realistic images after sampling from distribution as well as various applications including image inpainting and image-to-image translation.

This paper focuses on translating fully-furnished rooms into empty ones. Specifically, the task aims to hallucinate the occluded regions of an input image, thus after translation, yields an image from the same distribution but with a different context. In addition, the growing interest in AR/VR applications has increased the need for assisting applications in improving the user's experience. Concerning interior redecoration applications, diminishing objects from a scene is of paramount importance, a task that can be approached by explicitly translating the existing scene to a background-only scene. 

Moreover, $360^{\circ}$ devices get popularized, with multiple panorama datasets \cite{armeni2017joint,chang2017matterport3d,zheng2020structured3d} being available to facilitate the contemporary deep-learning-based methods. The wide field-of-view, provided by $360^{\circ}$ cameras further motivates the development of image synthesis approaches in the $360^{\circ}$ domain since it provides enough surrounding context information. 

Concerning occluded areas generation, the surrounding context must be rich to aid the generation process.   
In a sense, synthesizing occluded areas of a scene can be approached by image inpainting. Nonetheless, while image inpainting manages to generate plausible images that adhere to the distribution of the target image, it neglects to preserve the fidelity of the occluded structure. This demandingness makes the task lean toward an image-to-image translation problem. On the other hand, image-to-image translation methods aim to translate an image from one domain to another, with one of its applications being the transformation of semantic labels to real images. In that direction, owing to the need for both preserving the structure of the concealed region and the generation of plausible structures, \cite{gkitsas2021panodr} manage to approach the problem using a hybrid approach. However, this approach heavily depends on the necessity of a pre-trained dense layout network to predict the three dominant semantic classes of the scene(floor, wall, ceiling). This demand not only makes it hard to re-train or fine-tune the model on new datasets but also makes the training procedure complex and time-consuming. Further, the style modulation derived from \cite{zhu2020sean} does not take into account the hole-filling nature in the case of diminishing tasks. Specifically, the styles used for modulating the layer activations are not selected by considering the neighborhood of the hole but rather explicitly derive styles from the whole image. Finally, the single-stage approach is vulnerable to artifacts, principally in challenging scenarios.

In this work, we investigate empty room generation from fully-furnished ones, as a probe to comprehend the capability of neural networks to generate occluded regions of a scene. This is achieved by employing a two-stage, coarse-to-fine architecture.  To assist our network with the inferred scene structure, we overcome the semantic segmentation network requirement \cite{gkitsas2021panodr} by exploiting the feature space of the coarse network and train our model end-to-end. 
 
In summary, our contributions are:
\begin{itemize}
    \item We propose a full-to-empty room generation model that learns simultaneously to infer the scene dense layout, showing the benefits of end-to-end training for both training simplicity and model performance.
    \item Using an adapted semantic region adaptive normalization layer, we prove that we do not have to resort to non-differentiable semantic maps to modulate the layer activations.

    \end{itemize}


\begin{figure}
    \includegraphics[width=.47\textwidth]{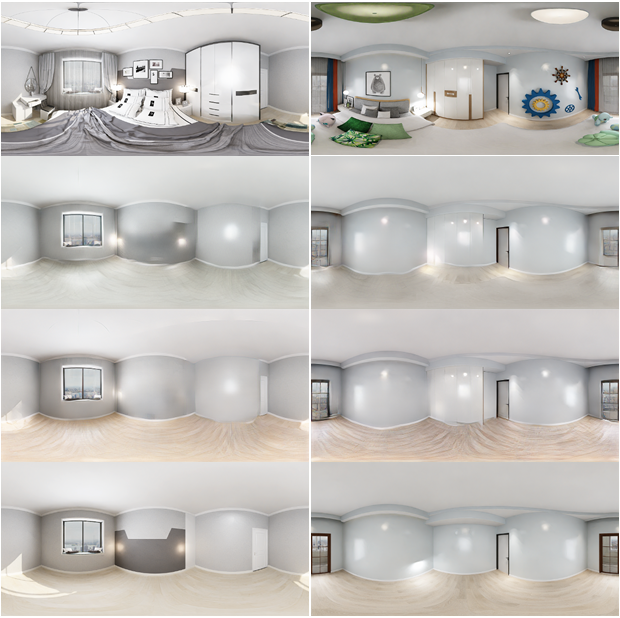}
    \captionof{figure}{Modeling the full-to-empty room generation task for spherical panoramas using our proposed approach. From top to bottom: The input panorama, fully-furnished, next the compared method \cite{gkitsas2021panodr} generated background panorama and the one predicted from our method, and finally the ground-truth. It is easily observable that our method produces more realistic results based on the ground truth empty panorama. By making the semantic region-adaptive normalization layer fully differentiable, the style modulation can effectively retain the style of the scene background even in extremely challenging cases.
     }
     \label{fig:teaser}
\end{figure}
\section{Related Work}
    
    \textbf{Image-to-image Translation:} Image-to-image translation approaches aim at translating a given source image to a corresponding image of a target domain. Over the last years, these approaches have gained increased attention, due to their applicability in a wide range of computer vision applications. Isola et al.\cite{isola2017image} first introduced the use of conditional GANs for tasks as translating semantic labels to images. In view of the recent works of conditional adversarial networks, SPADE \cite{park2019semantic} introduces the spatially-adaptive normalization layer in order to propel the semantic information provided by semantic masks in the deeper layers of the network. The modulation for the activations in normalization layers is accomplished via a spatial adaptive learned transformation. Accordingly, SEAN residual block  \cite{zhu2020sean} follows the same path while tackling the two shortcomings of SPADE. First, the use of only one style code to control the style of the whole generated image, and second, the absence of style contribution in the deeper layers of the network. Both drawbacks are alleviated by incorporating for each semantic class its corresponding style and thus using this style information via spatially varying normalization parameters.
    
    \textbf{Multi-Task Learning:} Deep learning multi-task methods aim at improving learned representation via simultaneously utilizing multiple learning-based tasks. Such approaches have been applied in several applications \cite{liu2015representation,jaderberg2016reinforcement}. An important application in this context is semantic segmentation. The incorporation of semantic segmentation task has been studied to perform detection or instance segmentation \cite{gidaris2015object,chen20153d,pinheiro2016learning}. Recently introduced, \cite{xu2021linear}, employs multi-task learning by leveraging the capability of generative networks to encode image semantics in its internal feature maps. Using a pre-trained GAN for generating an image from a latent vector, a simple linear transformation in the feature space is sufficient to provide the semantic segmentation map. To supervise the model, the ground truth semantic mask is obtained by a pre-trained semantic segmentation network while the standard cross-entropy loss is used as a loss function.  
    
    \textbf{Image Inpainting:} Traditional image inpainting methods fill the missing content by either searching the most similar patches in the background \cite{barnes2009patchmatch}
    or propagating neighboring structures \cite{sun2005image}. However, these approaches struggle when large regions are to be filled or content is unique and not present in the rest of the image. On the other hand, modern approaches \cite{iizuka2017globally,yu2018generative,yu2019free} leverage the recent advantages in deep learning to fill the missing regions by learning from a large corpus of data. More recently, some approaches assist the generation process by first estimating structural information such as edges \cite{nazeri2019edgeconnect} and edge-preserved smooth structures\cite{ren2019structureflow}. 
    
    \textbf{Diminished Reality:} Diminished Reality (DR) is the process of removing objects that are perceivable in our visual system. In order to diminish an object from a perceived view, background information is required. This prerequisites for the viewer to observe the occluded region from a different viewpoint or in advance \cite{mori2017survey}. Nonetheless, this is not feasible in cases where real-time requirements or technical substantial inability occur. In those cases, the occluded areas can be inferred by image inpainting approaches. \cite{gkitsas2021panodr} introduced a hybrid image inpainting, and image-to-image translation method to approach the DR problem. More specifically, first, the dense layout of the scene was inferred by a pre-trained semantic segmentation network. Next, the occluded regions were synthesized by a single-stage generator. Additionally, SEAN residual blocks \cite{zhu2020sean} were employed to modulate the normalized activations using the dense layout and the style codes from the input image.

\section{Approach}
\begin{figure*}[!htbp]
    \centering
    \includegraphics[width=0.85\textwidth]{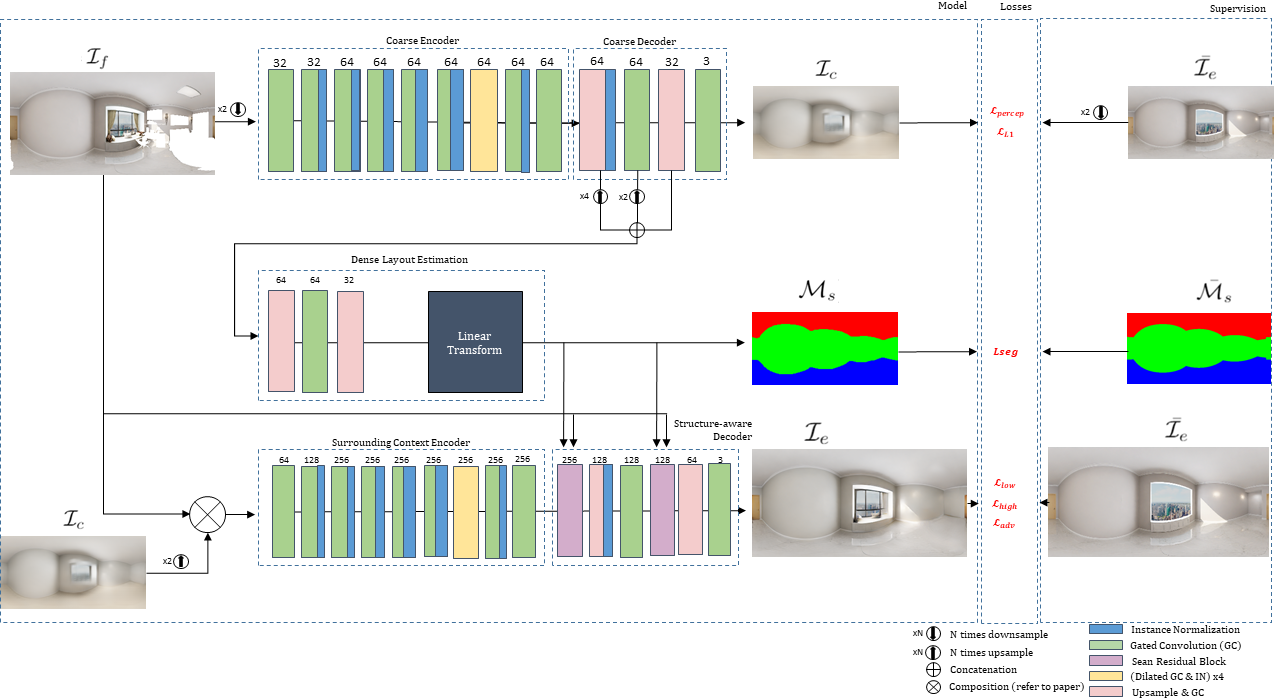}
    \caption{
    The architecture of the proposed method, along with the flow of data, supervision, and losses used in each stage of the end-to-end train. First, the input panorama is masked in the foreground area yielding $\im_{f}$ and fed to the coarse network. Next, after predicting the coarse background image $\im_{c}$, the dense layout of the scene $\mask_{s}$ is estimated by first applying upsampling and concatenation in its decoder feature space and afterward applying a linear transformation. Following, $\im_{c}$ is composited with $\im_{f}$ and fed to the refine network. After obtaining the encoded representation from the surrounding context encoder, $\mask{s}$ is used along with style code obtained from $\im_{f}$ for modulating the structure-aware decoder activations via SSEAN block. The generated background image, $\im_{c}$ is supervised by employing a low-level loss, a high-level loss as well an adversarial loss. The layout estimation is supervised by the standard cross-entropy loss. The architectures of the discriminator and SSEAN block are omitted for brevity.
    }
    \label{fig:network}
\end{figure*}

In this section we present our approach, which is depicted in Fig. \ref{fig:network} for modeling the translation from full to empty rooms.

First, we introduce the architecture of our generator, which is composed of a coarse and a refine network. Second, we propose a method for inferring the scene dense layout mask by leveraging the coarse network features. Next, we present an approach for overcoming the demand for a non-differentiable semantic mask as input in the SEAN residual block. Last but not least, regarding the diminishing task, we present a simple yet efficient method for enforcing styles close to holes to dominate in the generated occluded areas.

\subsection{Coarse Network}
Apart from the challenges posed by the need for hallucinating occluded areas, inferring such a region without first obtaining a coarse estimation is a challenging task. Albeit there exist one stage approaches for image inpainting \cite{zheng2019pluralistic,li2020recurrent} and diminished reality \cite{gkitsas2021panodr}, the quality of the generated content is prone to generate artifacts, especially when large holes occur. To mitigate this ambiguity we follow a two-stage coarse-to-fine architecture for our generator.
Firstly, the input image, of size $256 \times 512$, is down-sampled to resolution $128
\times 256$. Furthermore, we follow a slim architecture to reduce the parameters. The scope of the coarse network is to produce a coarse prediction, $\im_{c}$, that will be fed as input to the refine network in order to facilitate the hallucination of the occluded areas. In addition, the dense layout map of the scene $\mask_{s}$ is generated from the feature space of the decoder. Formally, given an input, furnished image, $\im_{f}$, with foreground objects to be removed, masked, we desire to learn a mapping for the coarse network, $G^{c}$, such that $$\{\im_{c},\mask_{s}\} = G^{c}(\im_{f}) $$

\subsection{Linear Transformation on Coarse Network Feature Space}
To provide the dense layout map of the scene, $\mask_{s}$  in the fine stage without necessitating a pre-trained model \cite{gkitsas2021panodr}, we follow a nuanced approach. 
More specifically, assuming the coarse generator $G^{c}$ is composed of an encoder $E^{c}$ and a decoder $D^{c}$, then $D^{c}$, in the $i^{th} layer$, comprises $x_{i}^{c} \in {\rm I\!R}^{h_{i} \times w_{i} \times c_{i}}$ feature maps. We denote with $h_{i},w_{i},c_{i}$, the height, width and channels of the $i^{th}$ layer, respectively. Additionally, we denote as $\feats \in {\rm I\!R}^{n \times h \times w}$ the upsampled feature maps, to the output image resolution, of the $D^{c}$, concatenated along the depth axis.

Given an input furnished image $\im_f$ to $E^{c}$, we seek to estimate a coarse prediction of the occluded background $\im_{c}$, alongside the dense layout of the scene. Our intuition for obtaining the latter stems from the hypothesis that the feature maps $\feats$ encode the semantics of the three abundant classes of an indoor scene(ceiling, wall, floor) and can be inferred by applying a linear transformation on $\feats$. 

Therefore, we aim to learn a mapping $F$ such that:
$$\mask_{s} = F(\feats)$$
Inspired by \cite{xu2021linear}, this mapping $F$ can be a linear transformation, defined as: 

$$\mask_{s} = \sum_{i=1}^{N-1} T_{i} (u_{i} \, x_{i}^{c}) = T \cdot \feats $$ where $u_{i}$ is the upsampling operation for the $i^{th}$ layer, and $T$  $\in {\rm I\!R}^{m \times n}$.

\subsection{Refine Network}
After obtaining the coarse estimation $\im_{c}$, we seek to eliminate generated artifacts from the coarse output via the refine stage. The architecture of the refine network $G^{r}$ is derived from \cite{gkitsas2021panodr}. In order to exploit the obtained $\mask_{s}$ from the coarse stage, we modify the blocks that leverage the dense layout map for modulating the normalized activations of the decoder.  


\subsection{Soft Semantic Region-Adaptive Normalization\label{sec:SSEAN}}

\begin{figure}
\begin{center}
    \includegraphics[width=1.00\columnwidth]{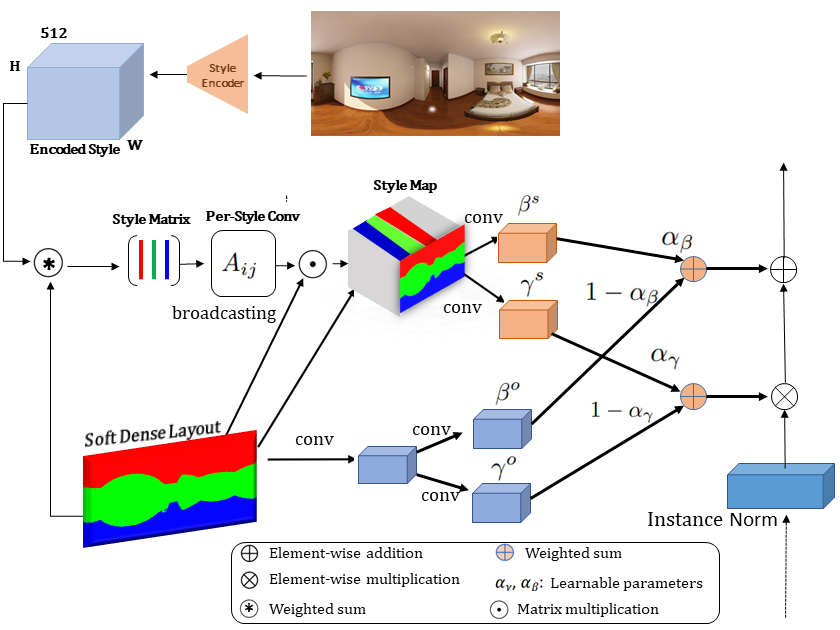}
    \captionof{figure}{Our Soft Semantic Region-Adaptive Normalization module which is built upon SEAN. First, the soft layout semantic mask is applied with a weighted sum on the encoded style, and afterward with a matrix multiplication with the styles obtained for each semantic region. For more info please refer to section \ref{sec:SSEAN}.}
    \label{fig:adapted_SEAN}

\end{center}
\end{figure}

In the recently introduced semantic region-adaptive normalization (SEAN) layer \cite{zhu2020sean}, the generation process is assisted by conditioning its output on one style code per semantic region. The original formulation in \cite{zhu2020sean} assumes one hot encoding for the semantic labels maps, which is typical in ground-truth annotated datasets. However, this demand hinders its applicability in end-to-end trainable models, due to the fact that the operations involved are not fully differentiable when the semantic label maps are provided by a predictive model, since in that case, the one-hot encoding is superseded by a probability distribution among possible semantic labels. In order to overcome this shortcoming, we make two adjustments in \cite{zhu2020sean}, the first related to the style encoder and the second related to the SEAN block.

In SEAN, the purpose of the style encoder is to encode one style code per semantic label. This is represented by a style matrix $\mathbf{ST} \in \mathbb{R}^{C\times D}$ where $C$ denotes the total number of possible semantic labels and $D$ the dimensionality of the style code. $\mathbf{ST}$ is computed via region-wise average pooling, from an intermediate matrix $\mathbf{\tilde{ST}} \in \mathbb{R}^{D\times H\times W}$ based on the semantic segmentation map $\mathbf{M} \in \mathbb{R}^{C\times H\times W}$, essentially averaging the style codes at all spatial locations for a given class. To make the process fully differentiable in cases where a pixel belongs to a class with a given probability, not necessarily strictly either zero or one (the previously mentioned one-hot encoding assumption), instead of using the one-hot encoded $\mathbf{M}$, we employ $\mask_s$. Let $\mathbf{\tilde{s}}_{i,j} \in \mathbb{R}^D$ denote the style code at the spatial location $(i,j)$ in $\mathbf{\tilde{ST}}$. Further, let $p_{i,j,c} \in [0,1]$ denote the probability that the pixel at spatial location $(i,j)$, belongs to class $c$ as defined by the value of the respective cell in $\mask_s$. Then, the style code $\mathbf{s}_c \in \mathbb{R}^D$ for the class $c$ is computed by the weighted sum:

\begin{equation}
    \mathbf{s}_c = \frac{1}{\sum_{i,j} p_{i,j,c}} \sum_{i,j} p_{i,j,c} \: \mathbf{\tilde{s}}_{i,j} 
\end{equation}

\noindent and the matrix $\mathbf{ST}$ is constructed by stacking $\{\mathbf{s}_c, \forall c\}$.

Regarding the SEAN block, we replace the broadcasting operation that produces the stylemap $\mathbf{SM} \in \mathbb{R}^{D\times H \times W}$ in a similar fashion. In particular, let $\mathbf{s'}_c \in \mathbb{R}^D$ denote the style code for class $c$ which is computed after the $1\times 1$ convolution with the respective style code in $\mathbf{ST}$. In this context, the purpose of the broadcast operation is to fill the stylemap's spatial locations at each pixel $(i,j)$ with the corresponding style code of the pixel's semantic label. The original broadcasting operation in SEAN does not take into account probability distributions other than one-hot. We make this operation soft and fully differentiable by assigning to each pixel of the stylemap the sum of all $\mathbf{s'}_c \forall c$ weighted by the pixel's probability to belong to class $c$. More formally, let $\mathbf{sm}_{i,j}$ denote the style code in style map spatial location $(i,j)$. Then:

\begin{equation}
\mathbf{sm}_{i,j} = \sum_c p_{i,j,c} \: \mathbf{s'}_c
\end{equation}

In both previously mentioned modifications in order to reduce the effect of mixing style codes belonging to different semantic labels, we pre-process the semantic label map $\mask_s$ (and consequently all $p_{i,j,c}$) via a sharpening operation powered by a softmax transformation parameterized by sharpening constant $K$:

\begin{equation}
    p'_{i,j,c} = \frac{e^{p_{i,j,c}/K}} {\sum_{c} e^{p_{i,j,c}/K}}
\end{equation}
In our experiments we empirically set $K=0.1$. To that end, we polarize $p_{i,j,c}$ towards the extreme values of $0$ and $1$. 

With those two modifications, we make SEAN fully differentiable and compatible with input segmentation masks following arbitrary probability distributions across semantic labels. Other than that, as depicted in Fig.\ref{fig:adapted_SEAN}, we keep the rest of the SEAN pipeline intact.

\subsection{Supervision}
In order to obtain the background image for a furnished on, we combine several losses to obtain  :

\vspace{-0.15cm}
\begin{equation}
    \loss = \loss_{low} + \loss_{high} + \loss_{adv} + \loss_{seg}.
    \vspace{-0.15cm}
\end{equation}
A low level reconstruction loss $\loss_{low}$, a high level synthesis loss $\loss_{high}$, an adaptive adversarial loss $\loss_{adv}$ and a layout estimation loss $\loss_{seg}$.

\textbf{Low-level Reconstruction Loss.}
This pixel-based loss focuses on the reconstruction of low frequency components of the predicted image $\im_e$:
\vspace{-0.15cm}

\begin{equation}
    \label{eq:reconstruction_loss}
    \begin{split}
    \loss_{low} = \lambda_{L1} \frac{1}{|\Omega|}||\mathbf{vec}(\Attention \odot |\bar{\im_{e}} - \im_{e}|)||_1 + \\
    \lambda_{TV} \bigg( \frac{1}{|\Omega_{\nabla_x}|}||\mathbf{vec}(\Attention \odot |\nabla_x\bar{\im_{e}}|\big)||_1 + \\
    \frac{1}{|\Omega_{\nabla_y}|}||\mathbf{vec}(\Attention \odot 
    |\nabla_y\bar{\im_{e}}|\big)||_1\bigg)
    \end{split}
\end{equation}

\noindent where $|\Omega|,|\Omega_{\nabla_x}|,|\Omega_{\nabla_y}|$ are the total number of pixels in $\bar{\im_e}$ and in the respective gradient images in $x$, $y$ directions. $\Attention \in \mathbb{R}^{W \times H}$ is the spherical attention mask used in \cite{zioulis2019spherical} that accounts for equirectangular distortion, while the $\mathbf{vec}$ operator treats its matrix argument as a flattened vector, and $|| \cdot ||_1$, denotes the L1-norm. Finally, $\odot$ denotes the Hadamard product.

Apart from the spherically weighted L1 loss, a total variation smoothness prior is used for the diminished area specifically to counter the high frequency artifacts usually seen in the early training stages of generative models.

\textbf{High-level Synthesis Loss.}
Apart from encouraging $\im_{e}$ and $\bar{\im_e}$ to have the same representation at the pixel level with $\loss_{low}$, we additionally employ a data-driven loss $\loss_{high}$.
This enforces them to have a similar representation in the feature space as computed by a CNN model $\vgg$, which in our case, is a pre-trained VGG-19 \cite{simonyan2014very}. 
Let $\vgg_{j}(\im) \in \mathbb{R}^{C_j\times H_j \times W_j}$ be the tensor of activations of the $j$-th layer of the network $\vgg$ with $C_j$ channels, for the given image $\im$, and $|\Omega_{\vgg_j}|$ the total number of elements of the tensor.

Then the loss is formulated as a combination of the perceptual and style losses:

\begin{align}
   \label{eq:synthesis_loss}    
   \loss_{high} &= \lambda_{perc} \loss_{perc} + \lambda_{style} \loss_{style}\\
   \loss_{perc} &= \sum_j \frac{1}{|\Omega_{{\vgg}_j}|}||\mathbf{vec}\big(\vgg_{j}(\bar{\im_{e}}) - \vgg_{j}(\im_{e})\big)||_1 \\
   \loss_{style} &= \sum_j \frac{1}{C_j} ||\mathbf{vec}\big(\pazocal{G}(\vgg_{j}(\bar{\im_{e}})) - \pazocal{G}(\vgg_{j}(\im_{e}))\big)||_1
\end{align}

\noindent $ \loss_{perc},\loss_{style}$ are the perceptual and style \cite{gatys2016image,johnson2016perceptual} losses, and $\pazocal{G}(\mathbf{M}) = \mathbf{M}\mathbf{M}^T, \mathbf{M} \in \mathbb{R}^{C_j \times (H_j\cdot W_j)}$ is the Gram matrix function.
Both losses are derived in a high dimensional data-driven feature space, with the former (perceptual) operating on a global level, and the latter (style) operating on global and local levels.

\textbf{Adaptive Adversarial Loss.}
To adaptively improve the quality of the generated background images $\im_e$ we additionally employ a discriminator-based loss that is learned during training.
Since we use a PatchGAN disciminator, we formulate our combined adversarial loss as a combination of a hinge loss on the final real/fake predictions \cite{lim2017geometric}, and a feature matching loss using the discriminator's intermediate features:
\begin{align}
\label{eq:adversarial_loss}
    \loss_{adv} &= \lambda_{D} \loss_{D} + \lambda_{FM} \loss_{FM} \\
    \loss_{D} &= \frac{1}{|\Omega_d|} \bigg(||\mathbf{vec}\big(r(1 - \mathbf{d}_e)\big)||_1 + ||\mathbf{vec}\big(r(1 + \mathbf{d}_{\hat{e}}\big)||_1\bigg)\\
    \loss_{FM} &= \sum_i \frac{1}{|\Omega_d^i|} ||\mathbf{vec}(\mathbf{d}_e^i - \mathbf{d}_{\hat{e}}^i)||_1,
\end{align}
where $\mathbf{d}_e$ and $\mathbf{d}_{\hat{e}}$ are the discriminator outputs for the real and predicted background images, $\Omega_d$ is the element domain of the discriminator's output, $|\Omega_d|$ the total count of its elements, while $i$ denotes intermediate discriminator feature maps and $|\Omega_d^i|$ their spatial element count. Finally, $r$ stands for the ReLU activation.
The spatial discriminator hinge loss and the feature matching loss are weighted by their respective weights.
Feature matching enforces the generator to minimize the statistical difference between the features of the ground truth images and the generated images, which helps further stabilize the training and improve the quality of the generated content. 

\textbf{Layout Estimation loss}: To supervise the dense layout estimation, we use 
the focal loss \cite{lin2017focal} which is proven to penalize the network better than the standard cross entropy loss on hard negative examples:
\begin{align}
\label{eq:focal_loss}
\loss_{seg} = a(1-p_{t})^{\gamma} \loss_{ce}(\bar{\mask_{s}}, \mask_{s})
\end{align}

With $p_{t}$ and $\loss_{ce}$ we denote the probabilities of the target class and the standard cross entropy loss, respectively. For our experiments we set $\alpha=0.25$ and $\gamma=2$.


\section{Results}
\textbf{Implementation details.}
We implement our model using PyTorch \cite{paszke2017automatic} with all experiments conducted on a Nvidia GeForce RTX 3090 GPU.
Our generative models are optimized using Adam \cite{kingma2014adam}, with $b_{1}=0.5$ and $b_{2}=0.999$, a learning rate of $0.0002$ and a batch size of $6$. 
The input and output panorama resolutions are $256 \times 128$ for the coarse network and $512 \times 256$ for the refine.
The weights the models are initialized from a zero-centered Normal distribution with $\sigma=0.02$. 
We empirically set $\lambda_{L1} = 4.0$, $\lambda_{TV} = 1.0$, $\lambda_{perc} = 0.15$, $\lambda_{style} = 40.0$, $\lambda_{D} = 0.2$ and $\lambda_{FM} = 20.0$.

\begin{figure*}[!htbp]
    \centering

\subfloat{\includegraphics[width=0.24\linewidth]{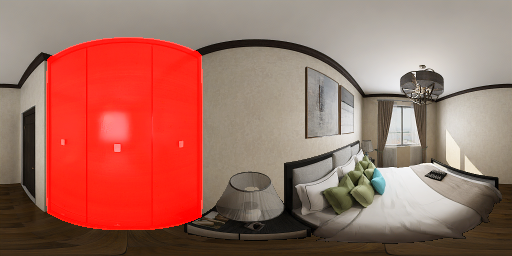}} 
\hfill
\subfloat{\includegraphics[width=0.24\linewidth]{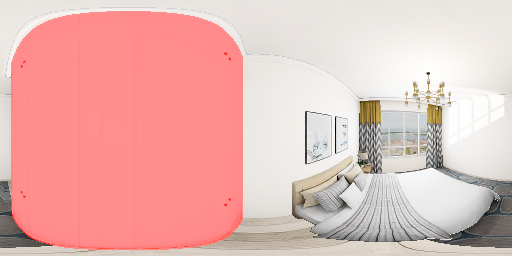}} 
\hfill
\subfloat{\includegraphics[width=0.24\linewidth]{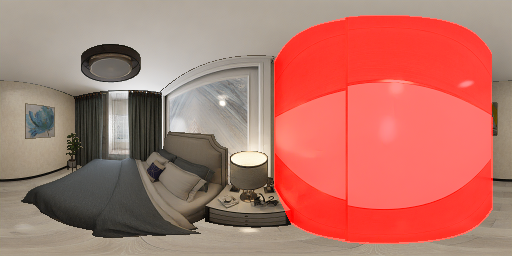}} 
\hfill
\subfloat{\includegraphics[width=0.24\linewidth]{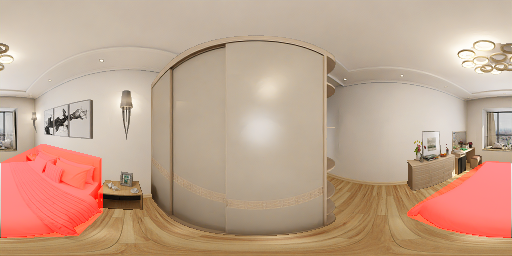}} 
\hfill

\subfloat{\includegraphics[width=0.24\linewidth]{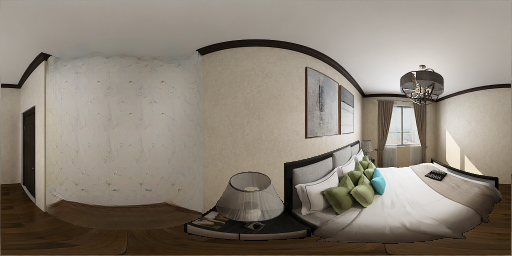}} 
\hfill
\subfloat{\includegraphics[width=0.24\linewidth]{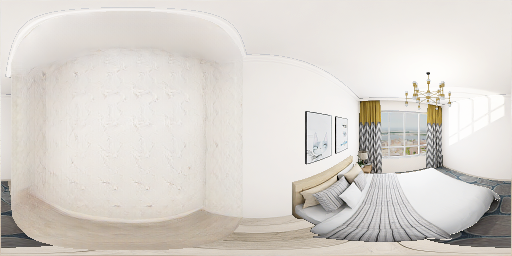}} 
\hfill
\subfloat{\includegraphics[width=0.24\linewidth]{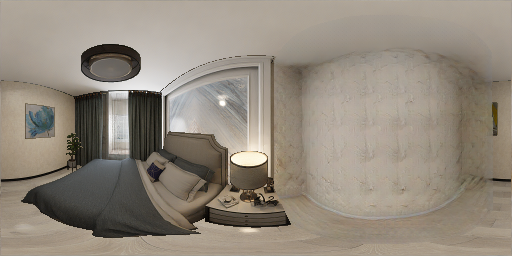}} 
\hfill
\subfloat{\includegraphics[width=0.24\linewidth]{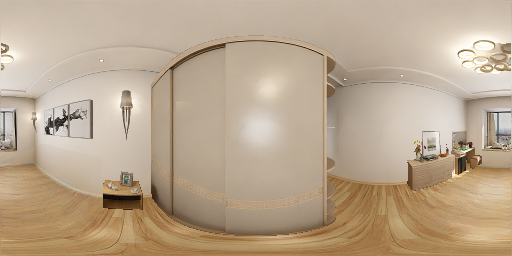}} 
\hfill

\hfill

\subfloat{\includegraphics[width=0.24\linewidth]{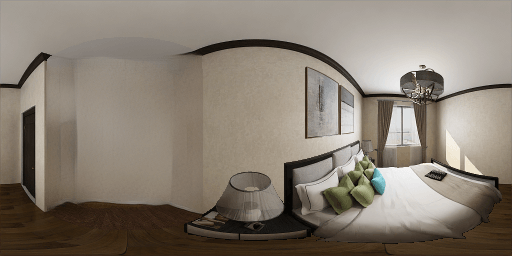}} 
\hfill
\subfloat{\includegraphics[width=0.24\linewidth]{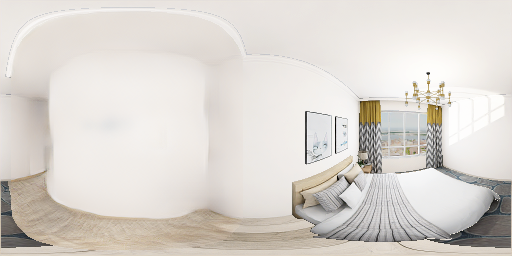}} 
\hfill
\subfloat{\includegraphics[width=0.24\linewidth]{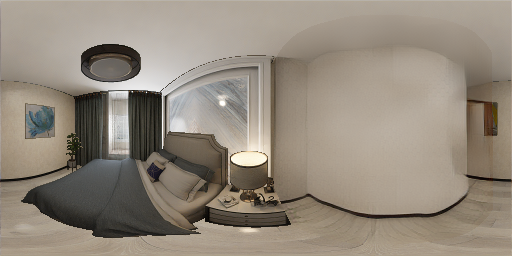}} 
\hfill
\subfloat{\includegraphics[width=0.24\linewidth]{figures/qualitative/PIC_scene_03492_2D_rendering_675223_panorama_out_0.png}} 
\hfill

\subfloat{\includegraphics[width=0.24\linewidth]{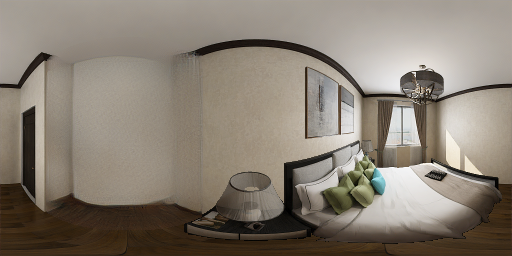}} 
\hfill
\subfloat{\includegraphics[width=0.24\linewidth]{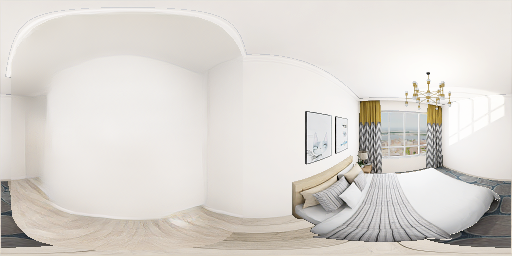}} 
\hfill
\subfloat{\includegraphics[width=0.24\linewidth]{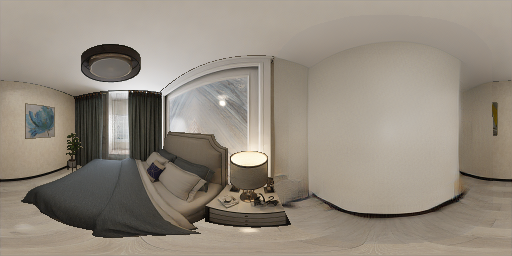}} 
\hfill
\subfloat{\includegraphics[width=0.24\linewidth]{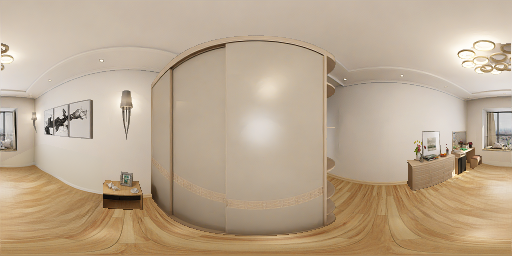}} 
\hfill

\caption{
Qualitative comparison on diminished reality application from scenes in our test set. From top to bottom: Input image with the diminished area masked with transparent red, PICNet, PanoDR, and ours. 
}
\label{fig:qualitative}
\end{figure*}



\textbf{Experiments.}
We compare our proposed method against the set of the current state-of-the art methods of PanoDR \cite{gkitsas2021panodr}, RFR\cite{li2020recurrent}, PICNet \cite{zheng2019pluralistic}. Moreover, to highlight the effectiveness of our method, we compare it against PanoDR-e2e. PanoDR-e2e is considered as the official work, but trained end-to-end, using our adapted SEAN residual block. All training configurations use the same adaptation of the Structured3D \cite{Structured3D} dataset as in \cite{gkitsas2021panodr}, with fixed seeds and using the official train/test splits of \cite{Structured3D}. The mentioned adaptation enables the applicability of the original dataset for diminished reality applications. For more details please refer to the original work of \cite{gkitsas2021panodr}.

\textbf{Quantitative comparisons.}
Table~\ref{tab:quantitative_results_diminishing} shows the performance of each method on the standard set of metrics, Mean Absolute Error (MAE), Peak to Signal-to-Noise Ratio (PSNR), Structural Similarity Index (SSIM), Fréchet inception distance (FID), and Learned Perceptual Image Patch Similarity (LPIPS), with the three first expressing similarity in a spatially local manner (pixel-wise or in small patches) and the latter two in a perceptual-global manner. More precisely, LPIPS compares the features extracted by a pre-trained VGG-16 model, rather than the images themselves, with the rationale being that the extracted features can be more expressive in local regions of the image. On the same page, FID compares the high-dimensional feature distributions of the predicted and ground-truth images, which assesses how close these distributions are across the dataset. Given the nature of our task, we aim to maximize the performance of the perceptual metrics, since preserving the the structure of the room is more visually appealing, rather than some minor photometric inconsistencies. Additionally, to assess the boundary preservation for the generated image, we follow the mIoU estimation for the introduced in \cite{gkitsas2021panodr}, by applying a pre-trained semantic segmentation network on the generated image and comparing it with the ground truth.

\begin{table*}[!htbp]
\centering
\caption{
Quantitative comparison on diminished reality application. Six metrics are used, FID, LPIPS, PSNR, SSIM, MAE and structural preservation (mIoU) on the Structured3D test set( $\downarrow$ means lower is better, and $\uparrow$ means higher is better).
}

\label{tab:quantitative_results_diminishing}
\begin{tabular}{@{}l|c|c|c|c|c | c | c}
\toprule
\textbf{Method}  & \textbf{FID $\downarrow$} &  \textbf{LPIPS $\downarrow$}  & \textbf{PSNR $\uparrow$} & \textbf{SSIM $\uparrow$} & \textbf{MAE $\downarrow$} & \textbf{mIoU $\uparrow$} \\ \midrule
RFR \cite{li2020recurrent} & 7.2474 & 0.0510 & 31.0114 & 0.9528 & 0.0067  & 0.8583     \\ \midrule
PICNet \cite{zheng2019pluralistic} & 6.7063  & 0.0533 & 32.3072 & 0.9557 & 0.0070  & 0.8502\\ \midrule
PanoDR \cite{gkitsas2021panodr} & 6.8374 &  0.0398 & 33.6611  & 0.9620 &\textbf{0.0058}  & 0.8768 \\ \midrule
PanoDR(e2e) & 7.2052 &  0.0357 & \textbf{33.6681}  & 0.9622 & 0.0060  & 0.8488 \\ \midrule
Ours & \textbf{6.6915} &  \textbf{0.0320} & 33.6576   & \textbf{0.9624} & \textbf{0.0058}  & \textbf{0.8789} \\ \midrule
\end{tabular}
\end{table*}








\textbf{Diminished reality application.}
Regarding the diminished reality application, given in Table \ref{tab:quantitative_results_diminishing}, it is apparent that our method not only surpasses the compared methods in terms of perceptual metrics (LPIPS, FID) but also exhibits equivalent performance concerning boundary preservation. More specifically, the model performance in terms of FID and LPIPS increases by 2.1\% and 16.6\% over the baseline, respectively whilst PSNR, SSIM, mIoU and MAE do not exhibit significant variations. This performance gain is attributed to the adapted SEAN residual block, which manages to handle in a better manner the style modulation of the scene. 

\textbf{Qualitative comparisons}
In order to further assess the quality of the model performance, we take a closer look at the qualitative comparisons. About the full-to-empty room generation, it can be observed in Fig.\ref{fig:teaser} that our method better preserves the overall hue of the scene than the compared method. In such challenging scenes where the objects that are to be diminished cover a large part of the scene, it is crucial the generated image preserves both structure and the hue that depict the scene. For instance, in the first row of Fig.~\ref{fig:teaser} the floor of the scene is almost covered by the bed and furniture of the room. PanoDR misses generating floor with the realistic visual result, while exhibits severe flaws with regard to the structure of the scene. Similarly, in the second row, PanoDR not only misses capturing the hue of the floor but also its output is blurry at the lower side of the scene. On the other hand, our method generates content with a hue not vastly different than of the ground truth.

Concerning the diminishing application, in Fig.\ref{fig:qualitative} we compare our method with PICNet, RFR, and PanoDR. The compared methods are prone to generating blurry images and in some cases, artifacts. In contrast, our approach achieves visually appealing results given the challenging nature of that cases. For example, one can see in the second column that the texture of the generated image is compatible with that of the surrounding context. In addition, albeit the object to be diminished covers almost the $40\%$ of the scene, its output is free of artifacts and blurry regions.  

\textbf{Monocular Depth Estimation task}

\begin{figure}[!htbp]
    \centering
    \includegraphics[width=0.49\textwidth]{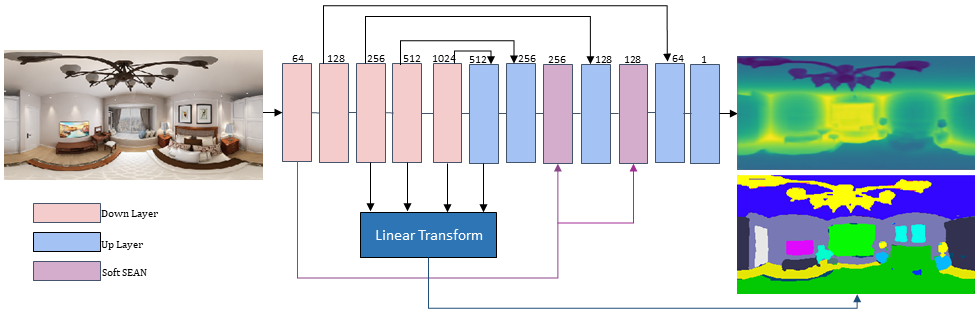}
    \caption{
    The architecture of the UNet adapted model using softSEAN block for depth estimation task. First, the input image is fed to the model, next, a linear transformation is applied to obtain the soft semantic segmentation of the scene, which is used to modulate the activation layers in the decoder.
    }
    \label{fig:depth_network}
\end{figure}

\begin{figure}[!htbp]
    \centering
    \subfloat{\includegraphics[width=0.24\linewidth]{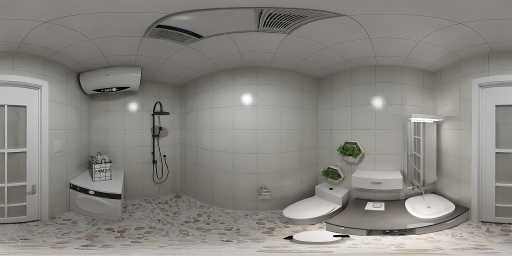}}
    \hfill
    \subfloat{\includegraphics[width=0.24\linewidth]{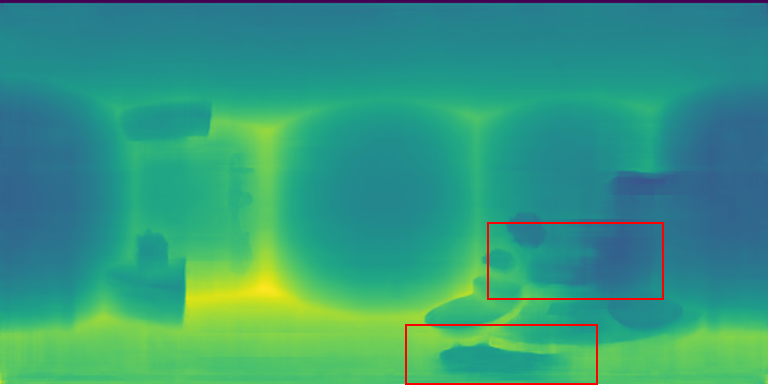}} 
    \hfill
    \subfloat{\includegraphics[width=0.24\linewidth]{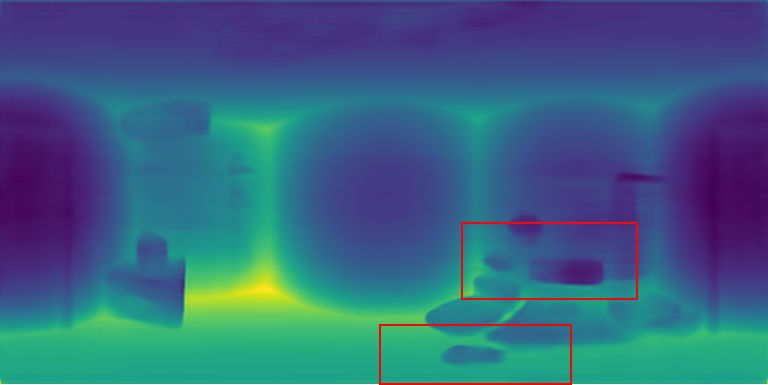}} 
    \hfill
    \subfloat{\includegraphics[width=0.24\linewidth]{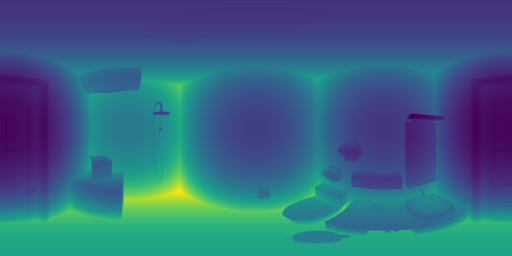}} 
    \hfill

    \subfloat{\includegraphics[width=0.24\linewidth]{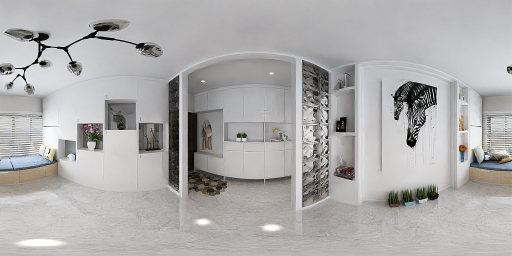}} 
    \hfill
    \subfloat{\includegraphics[width=0.24\linewidth]{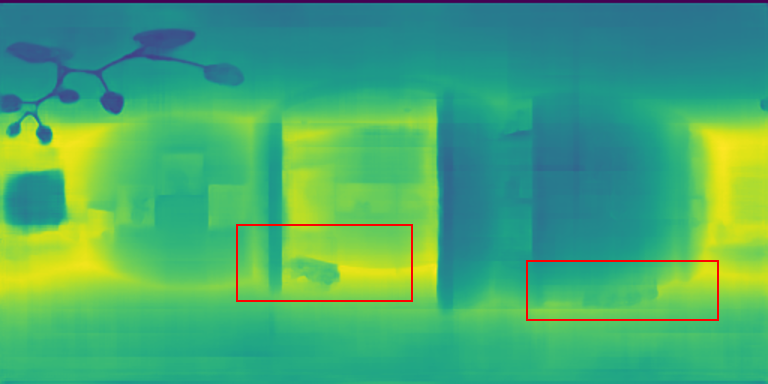}} 
    \hfill
    \subfloat{\includegraphics[width=0.24\linewidth]{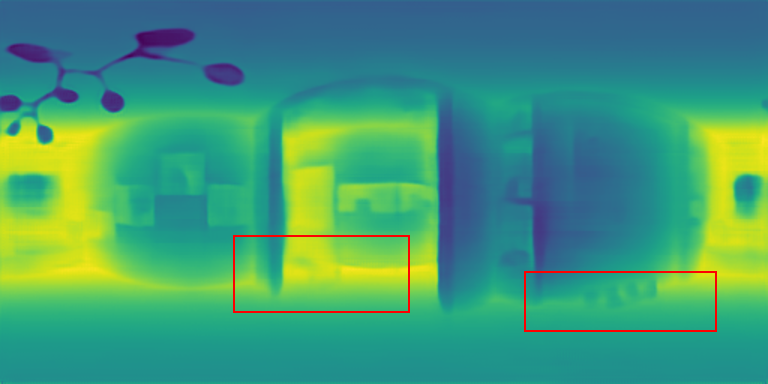}} 
    \hfill
    \subfloat{\includegraphics[width=0.24\linewidth]{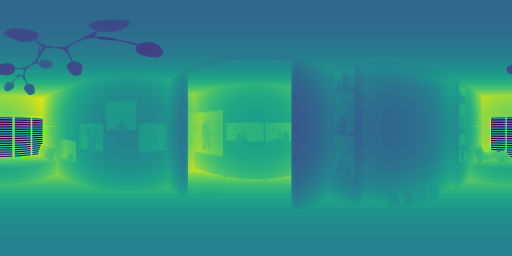}} 
    \hfill

    \subfloat{\includegraphics[width=0.24\linewidth]{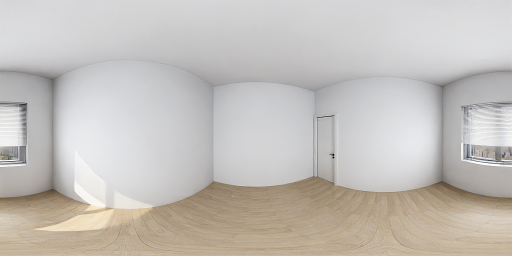}} 
    \hfill
    \subfloat{\includegraphics[width=0.24\linewidth]{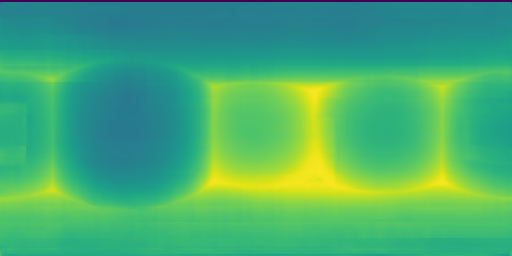}} 
    \hfill
    \subfloat{\includegraphics[width=0.24\linewidth]{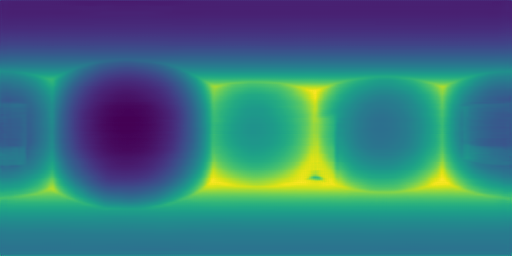}} 
    \hfill
    \subfloat{\includegraphics[width=0.24\linewidth]{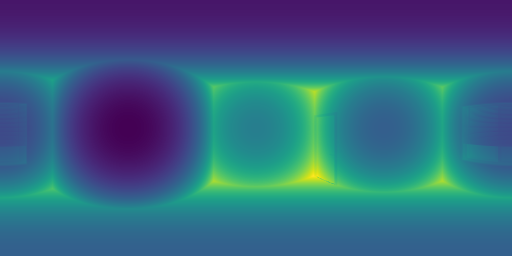}} 
    \hfill
    \caption{Qualitative results on omnidirectional dense depth estimation on samples of Structured3D. From left to right: input image, baseline, ours (Lt+softSEAN), ground truth.}
    \label{fig:depthEst}
\end{figure}

To further evaluate the softSEAN block, we conduct another experiment on a dense regression task, monocular depth estimation using the Structured3D dataset. We adapt UNet architecture \cite{ronneberger2015u} and derive the direct supervision from \cite{zioulis2019spherical}.
Similarly, we employ a linear transformation, $Lt$ to obtain the soft semantic mask of the input scene. Following a different line compared with the diminished reality task, we use the features from the second layer of the encoder to modulate the layers in the decoder. The main goal is to enhance the predictions for local regions through integrating the semantic map, composed of 41 classes. The network architecture is depicted in Fig.\ref{fig:depth_network}.
Quantitative results are presented in Table \ref{tab:depth_results},  for which we used typical metrics. The results enhance previous findings in that the softSEAN residual block along with the exploitation of the linear transformation block enforces the network to better predictions. Fig.\ref{fig:depthEst} clearly illustrates the benefits of our method, in terms of both local and global regions of the predicted depth map. For instance, in the second row, the baseline model misses predicting the correct depth due to texture transfer(highlighted on the left) while fails at capturing local objects' depth(highlighted on the right).  





\section{Conclusion}
In this work, we propose an approach for indoor spherical panoramas, in which an empty room is generated from a full-furnished one. The core idea of our method lies in using a two-stage coarse-to-fine network. First, a lightweight network is utilized to estimate a coarse prediction of the background while encodes in its features and generating the dense layout of the occluded regions of the scene. Subsequently, to leverage the latter for modulating layers activations, we adjust the SEAN block in a way that maintains the differentiability of the dense layout. Interestingly, we demonstrate that our method shows consistent improvement over the baselines regarding the diminished reality application while overcomes the barriers of previous methods and is trainable in an end-to-end manner. Further, we believe that the key insight of this work can be applied to room re-decoration and interior design applications. Last but not least, we validate the effectiveness of the soft SEAN residual block via applying it in the depth estimation task, showcasing its efficacy for different computer vision tasks. 
\\
\\

\begin{table}[!htbp]
\centering
\caption{
Results of omnidirectional depth estimation. The first row represents the baseline UNet architecture, the second the adapted model with linear transformation (Lt) and the latter the architecture that encompasses both linear transformation and softSEAN. 
}

\label{tab:depth_results}
\resizebox{\linewidth}{!}{%
\begin{tabular}{@{}l|c|c|c|c | c | c}
\toprule
\textbf{Method}   &  \textbf{RMSE $\downarrow$}  & \textbf{RMSE(log) $\downarrow$} & \textbf{$\delta1$ $\uparrow$} & \textbf{$\delta2$ $\uparrow$} & \textbf{$\delta3$ $\uparrow$} \\ \midrule
Baseline    & 0.4635 & 0.1738 & 0.9144 & 0.9613 & 0.9695  \\ \midrule
w $Lt$  & 0.4533 & 0.1684 & 0.9244 & 0.9675 & 0.9759  \\
\midrule
w $Lt$+$softSEAN$&\textbf{0.3820} & \textbf{0.1585}  & \textbf{0.9573} & \textbf{0.9768} & \textbf{0.9809} \\
\bottomrule
\end{tabular}
}
\end{table}

\textbf{Acknowledgements.}
This work was supported by the EC funded H2020 project ATLANTIS [GA 951900].

\bibliographystyle{apalike}
{\small
\bibliography{egbib}}

\end{document}